# AIS-BN: An Adaptive Importance Sampling Algorithm for Evidential Reasoning in Large Bayesian Networks

**Jian Cheng**                                                          JCHENG@SIS.PITT.EDU
**Marek J. Druzdzel**                                                    MAREK@SIS.PITT.EDU
*Decision Systems Laboratory*
*School of Information Sciences and Intelligent Systems Program*
*University of Pittsburgh, Pittsburgh, PA 15260 USA*

## Abstract

Stochastic sampling algorithms, while an attractive alternative to exact algorithms in very large Bayesian network models, have been observed to perform poorly in evidential reasoning with extremely unlikely evidence. To address this problem, we propose an adaptive importance sampling algorithm, AIS-BN, that shows promising convergence rates even under extreme conditions and seems to outperform the existing sampling algorithms consistently. Three sources of this performance improvement are (1) two heuristics for initialization of the importance function that are based on the theoretical properties of importance sampling in finite-dimensional integrals and the structural advantages of Bayesian networks, (2) a smooth learning method for the importance function, and (3) a dynamic weighting function for combining samples from different stages of the algorithm.

We tested the performance of the AIS-BN algorithm along with two state of the art general purpose sampling algorithms, likelihood weighting (Fung & Chang, 1989; Shachter & Peot, 1989) and self-importance sampling (Shachter & Peot, 1989). We used in our tests three large real Bayesian network models available to the scientific community: the CPCS network (Pradhan et al., 1994), the PATHFINDER network (Heckerman, Horvitz, & Nathwani, 1990), and the ANDES network (Conati, Gertner, VanLehn, & Druzdzel, 1997), with evidence as unlikely as $10^{-41}$. While the AIS-BN algorithm always performed better than the other two algorithms, in the majority of the test cases it achieved orders of magnitude improvement in precision of the results. Improvement in speed given a desired precision is even more dramatic, although we are unable to report numerical results here, as the other algorithms almost never achieved the precision reached even by the first few iterations of the AIS-BN algorithm.

## 1. Introduction

Bayesian networks (Pearl, 1988) are increasingly popular tools for modeling uncertainty in intelligent systems. With practical models reaching the size of several hundreds of variables (e.g., Pradhan et al., 1994; Conati et al., 1997), it becomes increasingly important to address the problem of feasibility of probabilistic inference. Even though several ingenious exact algorithms have been proposed, in very large models they all stumble on the theoretically demonstrated NP-hardness of inference (Cooper, 1990). The significance of this result can be observed in practice — exact algorithms applied to large, densely connected practical networks require either a prohibitive amount of memory or a prohibitive amount of computation and are unable to complete. While approximating inference to any desired precision has been shown to be NP-hard as well (Dagum & Luby, 1993), it is for very com-plex networks the only alternative that will produce any result at all. Furthermore, while obtaining the result is crucial in all applications, precision guarantees may not be critical for some types of problems and can be traded off against the speed of computation.

A prominent subclass of approximate algorithms is the family of stochastic sampling algorithms (also called *stochastic simulation* or *Monte Carlo* algorithms). The precision obtained by stochastic sampling generally increases with the number of samples generated and is fairly unaffected by the network size. Execution time is fairly independent of the topology of the network and is linear in the number of samples. Computation can be interrupted at any time, yielding an anytime property of the algorithms, important in time-critical applications.

While stochastic sampling performs very well in predictive inference, diagnostic reasoning, i.e., reasoning from observed evidence nodes to their ancestors in the network often exhibits poor convergence. When the number of observations increases, especially if these observations are unlikely a-priori, stochastic sampling often fails to converge to reasonable estimates of the posterior probabilities. Although this problem has been known since the very first sampling algorithm was proposed by Henrion (1988), little has been done to address it effectively. Furthermore, various sampling algorithms proposed were tested on simple and small networks, or networks with special topology, without the presence of extremely unlikely evidence and the practical significance of this problem has been underestimated. Given a typical number of samples used in real-time that are feasible on today's hardware, say $10^6$ samples, the behavior of a stochastic sampling algorithm will be drastically different for different size networks. While in a network consisting of 10 nodes and a few observations, it may be possible to converge to exact probabilities, in very large networks only a negligibly small fraction of the total sample space will be probed. One of the practical Bayesian network models that we used in our tests, a subset of the CPCS network (Pradhan et al., 1994), consists of 179 nodes. Its total sample space is larger than $10^{61}$. With $10^6$ samples, we can sample only $10^{-55}$ fraction of the sample space.

We believe that it is crucial (1) to study the feasibility and convergence properties of sampling algorithms on very large practical networks, and (2) to develop sampling algorithms that will show good convergence under extreme, yet practical conditions, such as evidential reasoning given extremely unlikely evidence. After all, small networks can be updated using any of the existing exact algorithms — it is precisely the very large networks where stochastic sampling can be most useful. As to the likelihood of evidence, we know that stochastic sampling will generally perform well when it is high (Henrion, 1988). So, it is important to look at those cases in which evidence is very unlikely. In this paper, we test two existing state of the art stochastic sampling algorithms for Bayesian networks, likelihood weighting (Fung & Chang, 1989; Shachter & Peot, 1989) and self-importance sampling (Shachter & Peot, 1989), on a subset of the CPCS network with extremely unlikely evidence. We show that they both exhibit similarly poor convergence rates. We propose a new sampling algorithm, that we call the *adaptive importance sampling* for Bayesian networks (AIS-BN), which is suitable for evidential reasoning in large multiply-connected Bayesian networks. The AIS-BN algorithm is based on importance sampling, which is a widely applied method for variance reduction in simulation that has also been applied in Bayesian networks (e.g., Shachter & Peot, 1989). We demonstrate empirically on three large practical Bayesian network models that the AIS-BN algorithm consistently outperforms the other two algorithms. In the majority of the test cases, it achieved over two orders of magnitude improvement in convergence. Improvement in speed given a desired precision is even more dramatic, although we are unable to report numerical results here, as the other algorithms never achieved the precision reached even by the first few iterations of the AIS-BN algorithm. The main sources of improvement are: (1) two heuristics for the initialization of the importance function that are based on the theoretical properties of importance sampling in finite-dimensional integrals and the structural advantages of Bayesian networks, (2) a smooth learning method for updating the importance function, and (3) a dynamic weighting function for combining samples from different stages of the algorithm. We study the value of the two heuristics used in the AIS-BN algorithm: (1) initialization of the probability distributions of parents of evidence nodes to uniform distribution and (2) adjusting very small probabilities in the conditional probability tables, and show that they both play an important role in the AIS-BN algorithm but only a moderate role in the existing algorithms.

The remainder of this paper is structured as follows. Section 2 first gives a general introduction to importance sampling in the domain of finite-dimensional integrals, where it was originally proposed. We show how importance sampling can be used to compute probabilities in Bayesian networks and how it can draw additional benefits from the graphical structure of the network. Then we develop a generalized sampling scheme that will aid us in reviewing the previously proposed sampling algorithms and in describing the AIS-BN algorithm. Section 3 describes the AIS-BN algorithm. We propose two heuristics for initialization of the importance function and discuss their theoretical foundations. We describe a smooth learning method for the importance function and a dynamic weighting function for combining samples from different stages of the algorithm. Section 4 describes the empirical evaluation of the AIS-BN algorithm. Finally, Section 5 suggests several possible improvements to the AIS-BN algorithm, possible applications of our learning scheme, and directions for future work.

## 2. Importance Sampling Algorithms for Bayesian Networks

We feel that it is useful to go back to the theoretical roots of importance sampling in order to be able to understand the source of speedup of the AIS-BN algorithm relative to the existing state of the art importance sampling algorithms for Bayesian networks. We first review the general idea of importance sampling in finite-dimensional integrals and how it can reduce the sampling variance. We then discuss the application of importance sampling to Bayesian networks. Readers interested in more details are directed to literature on Monte Carlo methods in computation of finite integrals, such as the excellent exposition by Rubinstein (1981) that we are essentially following in the first section.

### 2.1 Mathematical Foundations

Let $g(\mathbf{X})$ be a function of $m$ variables $\mathbf{X} = (X_1, ..., X_m)$ over a domain $\Omega \subset R^m$, such that computing $g(\mathbf{X})$ for any $\mathbf{X}$ is feasible. Consider the problem of approximate computation of the integral

$$I = \int_{\Omega} g(\mathbf{X}) \, d\mathbf{X} \; . \tag{1}$$

Importance sampling approaches this problem by writing the integral (1) as

$$I = \int_\Omega \frac{g(\mathbf{X})}{f(\mathbf{X})} f(\mathbf{X}) \, d\mathbf{X} \,,$$

where $f(\mathbf{X})$, often referred to as the *importance function*, is a probability density function over $\Omega$. $f(\mathbf{X})$ can be used in importance sampling if there exists an algorithm for generating samples from $f(\mathbf{X})$ and if the importance function is zero only when the original function is zero, i.e., $g(\mathbf{X}) \neq 0 \implies f(\mathbf{X}) \neq 0$.

After we have independently sampled $n$ points $\mathbf{s}_1, \mathbf{s}_2, \ldots, \mathbf{s}_n, \mathbf{s}_i \in \Omega$, according to the probability density function $f(\mathbf{X})$, we can estimate the integral $I$ by

$$\hat{I}_n = \frac{1}{n} \sum_{i=1}^n \frac{g(\mathbf{s}_i)}{f(\mathbf{s}_i)} \tag{2}$$

and estimate the variance of $\hat{I}_n$ by

$$\hat{\sigma}^2(\hat{I}_n) = \frac{1}{n \cdot (n-1)} \sum_{i=1}^n \left( \frac{g(\mathbf{s}_i)}{f(\mathbf{s}_i)} - \hat{I}_n \right)^2 \,. \tag{3}$$

It is straightforward to show that this estimator has the following properties:

1. $E(\hat{I}_n) = I$

2. $\lim_{n \to \infty} \hat{I}_n = I$

3. $\sqrt{n} \cdot (\hat{I}_n - I) \overset{n \to \infty}{\longrightarrow} \text{Normal}(0, \sigma_{f(\mathbf{X})}^2)$, where

$$\sigma_{f(\mathbf{X})}^2 = \int_\Omega \left( \frac{g(\mathbf{X})}{f(\mathbf{X})} - I \right)^2 f(\mathbf{X}) \, d\mathbf{X} \tag{4}$$

4. $E\left(\hat{\sigma}^2(\hat{I}_n)\right) = \sigma^2(\hat{I}_n) = \sigma_{f(\mathbf{X})}^2 / n$

The variance of $\hat{I}_n$ is proportional to $\sigma_{f(\mathbf{X})}^2$ and inversely proportional to the number of samples. To minimize the variance of $\hat{I}_n$, we can either increase the number of samples or try to decrease $\sigma_{f(\mathbf{X})}^2$. With respect to the latter, Rubinstein (1981) reports the following useful theorem and corollary.

**Theorem 1** *The minimum of $\sigma_{f(\mathbf{X})}^2$ is equal to*

$$\sigma_{f(\mathbf{X})}^2 = \left( \int_\Omega |g(\mathbf{X})| \, d\mathbf{X} \right)^2 - I^2$$

*and occurs when $\mathbf{X}$ is distributed according to the following probability density function*

$$f(\mathbf{X}) = \frac{|g(\mathbf{X})|}{\int_\Omega |g(\mathbf{X})| \, d\mathbf{X}} \,.$$

**Corollary 1** *If $g(\mathbf{X}) > 0$, then the optimal probability density function is*

$$f(\mathbf{X}) = \frac{g(\mathbf{X})}{I}$$

*and $\sigma^2_{f(\mathbf{X})} = 0$.*

Although in practice sampling from precisely $f(\mathbf{X}) = g(\mathbf{X})/I$ will occur rarely, we expect that functions that are close enough to it can still reduce the variance effectively. Usually, the closer the shape of the function $f(\mathbf{X})$ is to the shape of the function $g(\mathbf{X})$, the smaller is $\sigma^2_{f(\mathbf{X})}$. In high-dimensional integrals, selection of the importance function, $f(\mathbf{X})$, is far more critical than increasing the number of samples, since the former can dramatically affect $\sigma^2_{f(\mathbf{X})}$. It seems prudent to put more energy in choosing an importance function whose shape is as close as possible to that of $g(\mathbf{X})$ than to apply the brute force method of increasing the number of samples.

It is worth noting here that if $f(\mathbf{X})$ is uniform, importance sampling becomes a general Monte Carlo sampling. Another noteworthy property of importance sampling that can be derived from Equation 4 is that we should avoid $f(\mathbf{X}) \ll |g(\mathbf{X}) - I \cdot f(\mathbf{X})|$ in any part of the domain of sampling, even if $f(\mathbf{X})$ matches well $g(\mathbf{X})/I$ in important regions. If $f(\mathbf{X}) \ll |g(\mathbf{X}) - I \cdot f(\mathbf{X})|$, the variance can become very large or even infinite. We can avoid this by adjusting $f(\mathbf{X})$ to be larger in unimportant regions of the domain of $\mathbf{X}$.

While in this section we discussed importance sampling for continuous variables, the results stated are valid for discrete variables as well, in which case integration should be substituted by summation.

## 2.2 A Generic Importance Sampling Algorithm for Bayesian Networks

In the following discussion, all random variables used are multiple-valued, discrete variables. Capital letters, such as $A$, $B$, or $C$, denote random variables. Bold capital letters, such as $\mathbf{A}$, $\mathbf{B}$, or $\mathbf{C}$, denote sets of variables. Bold capital letter $\mathbf{E}$ will usually be used to denote the set of evidence variables. Lower case letters $a$, $b$, $c$ denote particular instantiations of variables $A$, $B$, and $C$ respectively. Bold lower case letters, such as $\mathbf{a}$, $\mathbf{b}$, $\mathbf{c}$, denote particular instantiations of sets $\mathbf{A}$, $\mathbf{B}$, and $\mathbf{C}$ respectively. Bold lower case letter $\mathbf{e}$, in particular, will be used to denote the observations, i.e., instantiations of the set of evidence variables $\mathbf{E}$. $Anc(A)$ denotes the set of ancestors of node $A$. $Pa(A)$ denotes the set of parents (direct ancestors) of node $A$. $pa(A)$ denotes a particular instantiation of $Pa(A)$. \ denotes set difference. $Pa(\mathbf{A})|_{\mathbf{E}=\mathbf{e}}$ denotes that we use the extended vertical bar to indicate substitution of $\mathbf{e}$ for $\mathbf{E}$ in $\mathbf{A}$.

We know that the joint probability distribution over all variables of a Bayesian network model, $\Pr(\mathbf{X})$, is the product of the probability distributions over each of the nodes conditional on their parents, i.e.,

$$\Pr(\mathbf{X}) = \prod_{i=1}^{n} \Pr(X_i | Pa(X_i)) \ . \tag{5}$$

In order to calculate $\Pr(\mathbf{E} = \mathbf{e})$, we need to sum over all $\Pr(\mathbf{X} \backslash \mathbf{E}, \mathbf{E} = \mathbf{e})$.

$$\Pr(\mathbf{E} = \mathbf{e}) = \sum_{\mathbf{X} \backslash \mathbf{E}} \Pr(\mathbf{X} \backslash \mathbf{E}, \mathbf{E} = \mathbf{e}) \tag{6}$$

We can see that Equation 6 is almost identical to Equation 1 except that integration is replaced by summation and the domain $\Omega$ is replaced by $\mathbf{X} \backslash \mathbf{E}$. The theoretical results derived for the importance sampling that we reviewed in the previous section can thus be directly applied to computing probabilities in Bayesian networks.

While there has been previous work on importance sampling-based algorithms for Bayesian networks, we will postpone the discussion of this work until the next section. Here we will present a generic stochastic sampling algorithm that will help us in both reviewing the prior work and in presenting our algorithm.

The posterior probability $\Pr(\mathbf{a}|\mathbf{e})$ can be obtained by first computing $\Pr(\mathbf{a}, \mathbf{e})$ and $\Pr(\mathbf{e})$ and then combining these based on the definition of conditional probability

$$\Pr(\mathbf{a}|\mathbf{e}) = \frac{\Pr(\mathbf{a}, \mathbf{e})}{\Pr(\mathbf{e})} \ . \tag{7}$$

In order to increase the accuracy of results of importance sampling in computing the posterior probabilities over different network variables given evidence, we should in general use different importance functions for $\Pr(\mathbf{a}, \mathbf{e})$ and for $\Pr(\mathbf{e})$. Doing so increases the computation time only linearly while the gain in accuracy may be significant given that obtaining a desired accuracy is exponential in nature. Very often, it is a common practice to use the same importance function (usually for $\Pr(\mathbf{e})$) to sample both probabilities. If the difference

---

1. Order the nodes according to their topological order.

2. Initialize importance function $\Pr^0(\mathbf{X} \backslash \mathbf{E})$, the desired number of samples $m$, the updating interval $l$, and the score arrays for every node.

3. $k \leftarrow 0$, $T \leftarrow \emptyset$

4. **for** $i \leftarrow 1$ **to** $m$ **do**

5.     **if** $(i \bmod l == 0)$ **then**

6.         $k \leftarrow k + 1$

7.         Update importance function $\Pr^k(\mathbf{X} \backslash \mathbf{E})$ based on $T$.
    **end if**

8.     $\mathbf{s}_i \leftarrow$ generate a sample according to $\Pr^k(\mathbf{X} \backslash \mathbf{E})$

9.     $T \leftarrow T \cup \{\mathbf{s}_i\}$

10.     Calculate Score$(\mathbf{s}_i, \Pr(\mathbf{X} \backslash \mathbf{E}, \mathbf{e}), \Pr^k(\mathbf{X} \backslash \mathbf{E}))$ and add it to the corresponding entry of every score array according to the instantiated states.
    **end for**

11. Normalize the score arrays for every node.

---

Figure 1: A generic importance sampling algorithm.

between the optimal importance functions for these two quantities is large, the performance may deteriorate significantly. Although $\widehat{\Pr}(\mathbf{a}, \mathbf{e})$ and $\widehat{\Pr}(\mathbf{e})$ are unbiased estimators according to Property 1 (Section 2.1), $\widehat{\Pr}(\mathbf{a}|\mathbf{e})$ obtained by means of Equation 7 is not an unbiased estimator. However, as the number of samples increases, the bias decreases and can be ignored altogether when the sample size is large enough (Fishman, 1995).

Figure 1 presents a generic stochastic sampling algorithm that captures most of the existing sampling algorithms. Without the loss of generality, we restrict ourselves in our description to so-called *forward sampling*, i.e., generation of samples in the topological order of the nodes in the network. The forward sampling order is accomplished by the initialization performed in Step 1, where parents of each node are placed before the node itself. In forward sampling, Step 8 of the algorithm, the actual generation of samples, works as follows. (*i*) each evidence node is instantiated to its observed state and is further omitted from sample generation; (*ii*) each root node is randomly instantiated to one of its possible states, according to the importance prior probability of this node, which can be derived from $\Pr^k(\mathbf{X}\backslash\mathbf{E})$; (*iii*) each node whose parents are instantiated is randomly instantiated to one of its possible states, according to the importance conditional probability distribution of this node given the values of the parents, which can also be derived from $\Pr^k(\mathbf{X}\backslash\mathbf{E})$; (*iv*) this procedure is followed until all nodes are instantiated. A complete instantiation $\mathbf{s}_i$ of the network based on this method is one sample of the joint importance probability distribution $\Pr^k(\mathbf{X}\backslash\mathbf{E})$ over all variables of the network. The scoring of Step 10 amounts to calculating $\Pr(\mathbf{s}_i, \mathbf{e})/\Pr^k(\mathbf{s}_i)$, as required by Equation 2. The ratio between the total score sum and the number of samples is an unbiased estimator of $\Pr(\mathbf{e})$. In Step 10, if we also count the score sum under the condition $\mathbf{A} = \mathbf{a}$, i.e., that some unobserved variables $\mathbf{A}$ have the values $\mathbf{a}$, the ratio between this score sum and the number of samples is an unbiased estimator of $\Pr(\mathbf{a}, \mathbf{e})$.

Most existing algorithms focus on the posterior probability distributions of individual nodes. As we mentioned above, for the sake of efficiency they count the score sum corresponding to $\Pr(A = a, \mathbf{e})$, $A \in \mathbf{X}\backslash\mathbf{E}$, and record it in an score array for node $A$. Each entry of this array corresponds to a specified state of $A$. This method introduces additional variance, as opposed to using the importance function derived from $\Pr^k(\mathbf{X}\backslash\mathbf{E})$ to sample $\Pr(A = a, \mathbf{e})$, $A \in \mathbf{X}\backslash\mathbf{E}$, directly.

## 2.3 Existing Importance Sampling Algorithms for Bayesian Networks

The main difference between various stochastic sampling algorithms is in how they process Steps 2, 7, and 8 in the generic importance sampling algorithm of Figure 1.

*Probabilistic logic sampling* (Henrion, 1988) is the simplest and the first proposed sampling algorithm for Bayesian networks. The importance function is initialized in Step 2 to $\Pr(\mathbf{X})$ and never updated (Step 7 is null). Without evidence, $\Pr(\mathbf{X})$ is the optimal importance function for the evidence set, which is empty anyway. It escapes most authors that $\Pr(\mathbf{X})$ may be not the optimal importance function for $\Pr(A = a)$, $A \in \mathbf{X}$, when $A$ is not a root node. A mismatch between the optimal and the actually used importance function may result in a large variance. The sampling process with evidence is the same as without evidence except that in Step 10 we do not count the scores for those samples that are inconsistent with the observed evidence, which amounts to discarding them. When the evidence is very unlikely, there is a large difference between $\Pr(\mathbf{X})$ and the optimal importance function. Effectively, most samples are discarded and the performance of logic sampling deteriorates badly.

*Likelihood weighting* (LW) (Fung & Chang, 1989; Shachter & Peot, 1989) enhances the logic sampling in that it never discards samples. In likelihood weighting, the importance function in Step 2 is

$$\Pr(\mathbf{X}\backslash\mathbf{E}) = \prod_{x_i \notin \mathbf{e}} \Pr(x_i | \mathrm{Pa}(X_i)) \bigg|_{\mathbf{E}=\mathbf{e}} .$$

Likelihood weighting does not update the importance function in Step 7. Although likelihood weighting is an improvement on logic sampling, its convergence rate can be still very slow when there is large difference between the optimal importance function and $\Pr(\mathbf{X}\backslash\mathbf{E})$, again especially in situations when evidence is very unlikely. Because of its simplicity, the likelihood weighting algorithm has been the most commonly used simulation method for Bayesian network inference. It often matches the performance of other, more sophisticated schemes because it is simple and able to increase its precision by generating more samples than other algorithms in the same amount of time.

*Backward sampling* (Fung & del Favero, 1994) changes Step 1 of our generic algorithm and allows for generating samples from evidence nodes in the direction that is opposite to the topological order of nodes in the network. In Step 2, backward sampling uses the likelihood of some of the observed evidence and some instantiated nodes to calculate $\Pr^0(\mathbf{X}\backslash\mathbf{E})$. Although Fung and del Favero mentioned the possibility of dynamic node ordering, they did not propose any scheme for updating the importance function in Step 7. Backward sampling suffers from problems that are similar to those of likelihood weighting, i.e., a possible mismatch between its importance function and the optimal importance function can lead to poor convergence.

*Importance sampling* (Shachter & Peot, 1989) is the same as our generic sampling algorithm. Shachter and Peot introduced two variants of importance sampling: *self-importance* (SIS) and *heuristic importance*. The importance function used in the first step of the self-importance algorithm is

$$\Pr^0(\mathbf{X}\backslash\mathbf{E}) = \prod_{x_i \notin \mathbf{e}} \Pr(x_i | \mathrm{Pa}(X_i)) \bigg|_{\mathbf{E}=\mathbf{e}} .$$

This function is updated in Step 7. The algorithm tries to revise the conditional probability tables (CPTs) periodically in order to make the sampling distribution gradually approach the posterior distribution. Since the same data are used to update the importance function and to compute the estimator, this process introduces bias in the estimator. Heuristic importance first removes edges from the network until it becomes a polytree, and then uses a modified version of the polytree algorithm (Pearl, 1986) to compute the likelihood functions for each of the unobserved nodes. $\Pr^0(\mathbf{X}\backslash\mathbf{E})$ is a combination of these likelihood functions with $\Pr(\mathbf{X}\backslash\mathbf{E},\mathbf{e})$. In Step 7 heuristic importance does not update $\Pr^k(\mathbf{X}\backslash\mathbf{E})$. As Shachter and Peot (1989) point out, this heuristic importance function can still lead to a bad approximation of the optimal importance function. There exist also other algorithms such as a combination of self-importance and heuristic importance (Shachter & Peot, 1989;

Shwe & Cooper, 1991). Although some researchers suggested that this may be a promising direction for the work on sampling algorithms, we have not seen any results that would follow up on this.

A separate group of stochastic sampling methods is formed by so-called *Markov Chain Monte Carlo (MCMC)* methods that are divided into Gibbs sampling, Metropolis sampling, and Hybrid Monte Carlo sampling (Geman & Geman, 1984; Gilks, Richardson, & Spiegelhalter, 1996; MacKay, 1998). Roughly speaking, these methods draw random samples from an unknown target distribution $f(\mathbf{X})$ by biasing the search for this distribution towards higher probability regions. When applied to Bayesian networks (Pearl, 1987; Chavez & Cooper, 1990) this approach determines the sampling distribution of a variable from its previous sample given its Markov blanket (Pearl, 1988). This corresponds to updating $\Pr^k(\mathbf{X}\backslash\mathbf{E})$ when sampling every node. $\Pr^k(\mathbf{X}\backslash\mathbf{E})$ will converge to the optimal importance function for $\Pr(\mathbf{e})$ if $\Pr^0(\mathbf{X}\backslash\mathbf{E})$ satisfies some ergodic properties (York, 1992). Since the convergence to the limiting distribution is very slow and calculating updates of the sampling distribution is costly, these algorithms are not used in practice as often as the simple likelihood weighting scheme.

There are also some other simulation algorithms, such as bounded variance algorithm (Dagum & Luby, 1997) and the AA algorithm (Dagum et al., 1995), which are essentially based on the LW algorithm and the Stopping-Rule Theorem (Dagum et al., 1995). Cano et al. (1996) proposed another importance sampling algorithm that performed somewhat better than LW in cases with extreme probability distributions, but, as the authors state, in general cases it "produced similar results to the likelihood weighting algorithm." Hernandez et al. (1998) also applied importance sampling and reported a moderate improvement on likelihood weighting.

## 2.4 Practical Performance of the Existing Sampling Algorithms

The largest network that has been tested using sampling algorithms is QMR-DT (Quick Medical Reference — Decision Theoretic) (Shwe et al., 1991; Shwe & Cooper, 1991), which contains 534 adult diseases and 4,040 findings, with 40,740 arcs depicting disease-to-finding dependencies. The QMR-DT network belongs to a class of special bipartite networks and its structure is often referred to as BN2O (Henrion, 1991), because of its two-layer composition: disease nodes in the top layer and finding nodes in the bottom layer. Shwe and colleagues used an algorithm combining self-importance and heuristic importance and tested its convergence properties on the QMR-DT network. But since the heuristic method *iterative tabular Bayes* (ITB) that makes use of a version of Bayes' rule is designed for the BN2O networks, it cannot be generalized to arbitrary networks. Although Shwe and colleagues concluded that Markov blanket scoring and self-importance sampling significantly improve the convergence rate in their model, we cannot extend this conclusion to general networks. The computation of Markov blanket scoring is more complex in a general multi-connected network than in a BN2O network. Also, the experiments conducted lacked a gold-standard posterior probability distribution that could serve to judge the convergence rate.

Pradhan and Dagum (1996) tested an efficient version of the LW algorithm — bounded variance algorithm (Dagum & Luby, 1997) and the AA algorithm (Dagum et al., 1995) on a 146 node, multiply connected medical diagnostic Bayesian network. One limitation in their tests is that the probability of evidence in the cases selected for testing was rather high. Although over 10% of the cases had the probability of evidence on the order of $10^{-8}$ or smaller, a simple calculation based on the reported mean $\mu = 34.5$ number of evidence nodes, shows that the average probability of an observed state of an evidence node conditional on its direct predecessors was on the order of $(10^{-8})^{1/34.5} \approx 0.59$. Given that their algorithm is essentially based on the LW algorithm, based on our tests we suspect that the performance will deteriorate on cases where the evidence is very unlikely. Both algorithms focus on the marginal probability of one hypothesis node. If there are many queried nodes, the efficiency may deteriorate.

We have tested the algorithms discussed in Section 2.3 on several large networks. Our experimental results show that in cases with very unlikely evidence, none of these algorithms converges to reasonable estimates of the posterior probabilities within a reasonable amount of time. The convergence becomes worse as the number of evidence nodes increases. Thus, when using these algorithms in very large networks, we simply cannot trust the results. We will present results of tests of the LW and SIS algorithms in more detail in Section 4.

## 3. AIS-BN: Adaptive Importance Sampling for Bayesian Networks

The main reason why the existing stochastic sampling algorithms converge so slowly is that they fail to learn a good importance function during the sampling process and, effectively, fail to reduce the sampling variance. When the importance function is optimal, such as in probabilistic logic sampling without any evidence, each of the algorithms is capable of converging to fairly good estimates of the posterior probabilities within relatively few samples. For example, assuming that the posterior probabilities are not extreme (i.e., larger than say 0.01), as few as 1,000 samples may be sufficient to obtain good estimates. In this section, we present the adaptive importance sampling algorithm for Bayesian networks (AIS-BN) that, as we will demonstrate in the next section, performs very well on most tests. We will first describe the details of the algorithm and prove two theorems that are useful in learning the optimal importance sampling function.

### 3.1 Basic Algorithm — AIS-BN

Compared with importance sampling used in normal finite-dimensional integrals, importance sampling used in Bayesian networks has several significant advantages. First, the network joint probability distribution $\Pr(\mathbf{X})$ is decomposable and can be factored into component parts. Second, the network has a clear structure, which represents many conditional independence relationships. These properties are very helpful in estimating the optimal importance function.

The basic AIS-BN algorithm is presented in Figure 2. The main differences between the AIS-BN algorithm and the basic importance sampling algorithm in Figure 1 is that we introduce a monotonically increasing weight function $w^k$ and two effective heuristic initialization methods in Step 2. We also introduce a special learning component in Step 7 to let the updating process run more smoothly, avoiding oscillation of the parameters. The

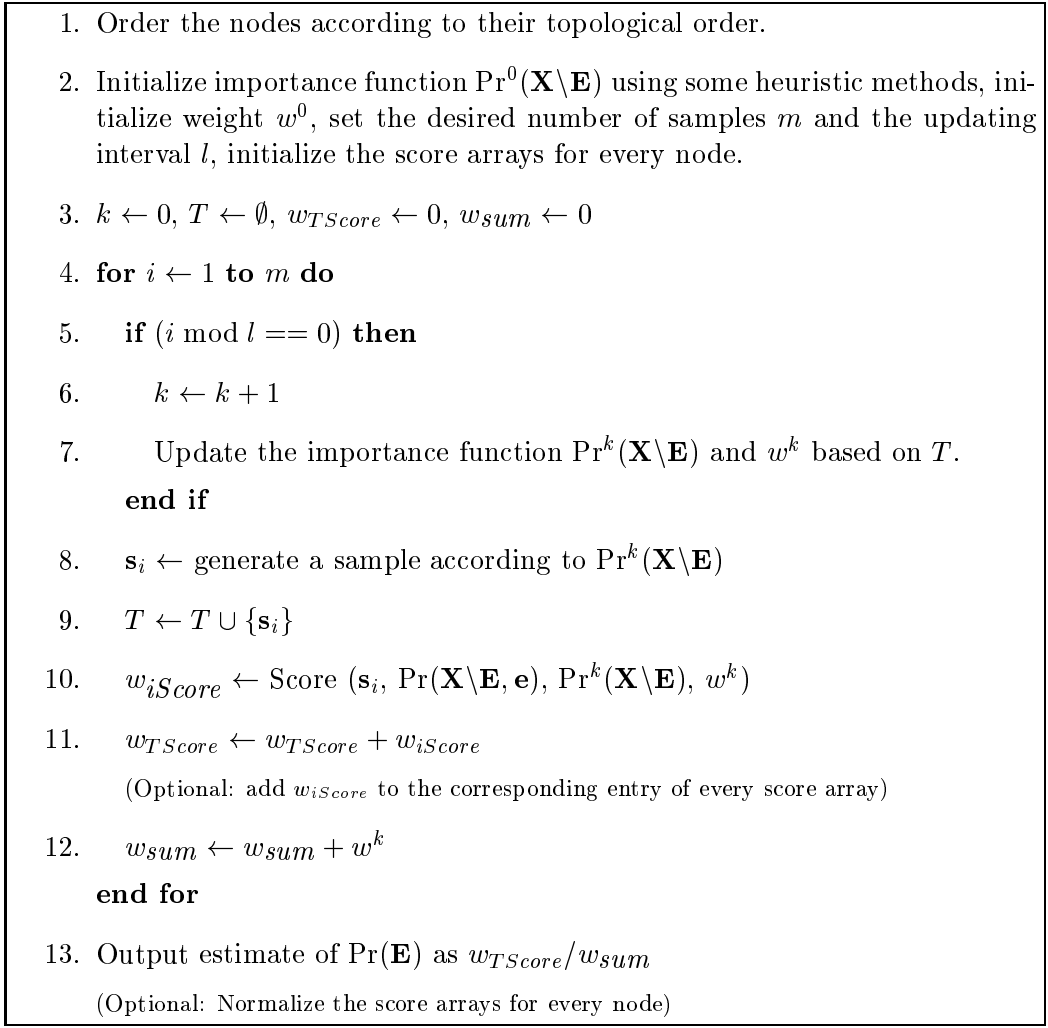

1. Order the nodes according to their topological order.

2. Initialize importance function $\Pr^0(\mathbf{X}\backslash\mathbf{E})$ using some heuristic methods, initialize weight $w^0$, set the desired number of samples $m$ and the updating interval $l$, initialize the score arrays for every node.

3. $k \leftarrow 0$, $T \leftarrow \emptyset$, $w_{TScore} \leftarrow 0$, $w_{sum} \leftarrow 0$

4. **for** $i \leftarrow 1$ **to** $m$ **do**

5.    **if** $(i \bmod l == 0)$ **then**

6.       $k \leftarrow k + 1$

7.       Update the importance function $\Pr^k(\mathbf{X}\backslash\mathbf{E})$ and $w^k$ based on $T$.
   **end if**

8.    $\mathbf{s}_i \leftarrow$ generate a sample according to $\Pr^k(\mathbf{X}\backslash\mathbf{E})$

9.    $T \leftarrow T \cup \{\mathbf{s}_i\}$

10.    $w_{iScore} \leftarrow$ Score $(\mathbf{s}_i, \Pr(\mathbf{X}\backslash\mathbf{E}, \mathbf{e}), \Pr^k(\mathbf{X}\backslash\mathbf{E}), w^k)$

11.    $w_{TScore} \leftarrow w_{TScore} + w_{iScore}$

     (Optional: add $w_{iScore}$ to the corresponding entry of every score array)

12.    $w_{sum} \leftarrow w_{sum} + w^k$
   **end for**

13. Output estimate of $\Pr(\mathbf{E})$ as $w_{TScore}/w_{sum}$

   (Optional: Normalize the score arrays for every node)

Figure 2: The adaptive importance sampling for Bayesian Networks (AIS-BN) algorithm.

score processing in Step 10 is

$$w_{iScore} = w^k \frac{\Pr(\mathbf{s}_i, \mathbf{e})}{\Pr^k(\mathbf{s}_i)}\,.$$

Note that in this respect the algorithm in Figure 1 becomes a special case of AIS-BN when $w^k = 1$. The reason why we use $w^k$ is that we want to give different weights to the sampling results obtained at different stages of the algorithm. As each stage updates the importance function, they will all have different distance from the optimal importance function. We recommend that $w^k \propto 1/\hat{\sigma}^k$, where $\hat{\sigma}^k$ is the standard deviation estimated in stage $k$ using Equation 3.[1] In order to keep $w^k$ monotonically increasing, if $w^k$ is smaller than $w^{k-1}$, we adjust its value to $w^{k-1}$. This weighting scheme may introduce bias into the final result. Since the initial importance sampling functions are often inefficient and introduce big variance into the results, we also recommend that $w^k = 0$ in the first few stages of the algorithm. We have designed this weighting scheme to reflect the fact that in practice estimates with very small estimated variance are usually good estimates.

---

1. A similar weighting scheme based on variance was apparently developed independently by Ortiz and Kaelbling (2000), who recommend the weight $w^k \propto 1/(\hat{\sigma}^k)^2$.

## 3.2 Modifying the Sampling Distribution in AIS-BN

Based on the theoretical considerations of Section 2.1, we know that the crucial element of the algorithm is converging on a good approximation of the optimal importance function. In what follows, we first give the optimal importance function for calculating $\Pr(\mathbf{E} = \mathbf{e})$ and then discuss how to use the structural advantages of Bayesian networks to approximate this function. In the sequel, we will use the symbol $\rho$ to denote the importance sampling function and $\rho^*$ to denote the optimal importance sampling function.

Since $\Pr(\mathbf{X} \backslash \mathbf{E}, \mathbf{E} = \mathbf{e}) > 0$, from Corollary 1 we have

$$\rho(\mathbf{X} \backslash \mathbf{E}) = \frac{\Pr(\mathbf{X} \backslash \mathbf{E}, \mathbf{E} = \mathbf{e})}{\Pr(\mathbf{E} = \mathbf{e})} = \Pr(\mathbf{X} | \mathbf{E} = \mathbf{e}) \ .$$

The following corollary captures this result.

**Corollary 2** *The optimal importance sampling function $\rho^*(\mathbf{X} \backslash \mathbf{E})$ for calculating $\Pr(\mathbf{E} = \mathbf{e})$ in Equation 6 is $\Pr(\mathbf{X} | \mathbf{E} = \mathbf{e})$.*

Although we know the mathematical expression for the optimal importance sampling function, it is difficult to obtain this function exactly. In our algorithm, we use the following importance sampling function

$$\rho(\mathbf{X} \backslash \mathbf{E}) = \prod_{i=1}^{n} \Pr(X_i | \mathrm{Pa}(X_i), \mathbf{E}) \ . \tag{8}$$

This function partially considers the effect of all the evidence on every node during the sampling process. When the network structure is the same as that of the network which has absorbed the evidence, this function is the optimal importance sampling function. It is easy to learn and, as our experimental results show, it is a good approximation to the optimal importance sampling function. Theoretically, when the posterior structure of the model changes drastically as the result of observed evidence, this importance sampling function may perform poorly. We have tried to find practical networks where this would happen, but to the day have not encountered a drastic example of this effect.

From Section 2.2, we know that the score sums corresponding to $\{x_i, \mathrm{pa}(X_i), \mathbf{e}\}$ can yield an unbiased estimator of $\Pr(x_i, \mathrm{pa}(X_i), \mathbf{e})$. According to the definition of conditional probability, we can get an estimator of $\Pr'(x_i | \mathrm{pa}(X_i), \mathbf{e})$. This can be achieved by maintaining an updating table for every node, the structure of which mimics the structure of the CPT. Such tables allow us to decompose the above importance function into components that can be learned individually. We will call these tables the *importance conditional probability tables* (ICPT).

**Definition 1** *An* importance conditional probability table (ICPT) *of a node $X$ is a table of posterior probabilities $\Pr(X | \mathrm{Pa}(X), \mathbf{E} = \mathbf{e})$ conditional on the evidence and indexed by its immediate predecessors,* $\mathrm{Pa}(X)$.

The ICPT tables will be modified during the process of learning the importance function. Now we will prove a useful theorem that will lead to considerable savings in the learning process.

**Theorem 2**

$$X_i \in \mathbf{X}, X_i \notin \mathrm{Anc}(\mathbf{E}) \Rightarrow \Pr(X_i|\mathrm{Pa}(X_i), \mathbf{E}) = \Pr(X_i|\mathrm{Pa}(X_i)) . \tag{9}$$

**Proof:** Suppose we have set the values of all the parents of node $X_i$ to $\mathrm{pa}(X_i)$. Node $X_i$ is dependent on evidence $\mathbf{E}$ given $\mathrm{pa}(X_i)$ only when $X_i$ is $d$-connecting with $\mathbf{E}$ given $\mathrm{pa}(X_i)$ (Pearl, 1988). According to the definition of $d$-connectivity, this happens only when there exists a member of $X_i$'s descendants that belongs to the set of evidence nodes $\mathbf{E}$. In other words $X_i \notin \mathrm{Anc}(\mathbf{E})$. □

Theorem 2 is very important for the AIS-BN algorithm. It states essentially that the ICPT tables of those nodes that are not ancestors of the evidence nodes are equal to the CPT tables throughout the learning process. We only need to learn the ICPT tables for the ancestors of the evidence nodes. Very often this can lead to significant savings in computation. If, for example, all evidence nodes are root nodes, we have our ICPT tables for every node already and the AIS-BN algorithm becomes identical to the likelihood weighting algorithm. Without evidence, the AIS-BN algorithm becomes identical to the probabilistic logic sampling algorithm.

It is worth pointing out that for some $X_i$, $\Pr(X_i|P_a(X_i), \mathbf{E})$ (i.e., the ICPT table for $X_i$), can be easily calculated using exact methods. For example, when $X_i$ is the only parent of an evidence node $E_j$ and $E_j$ is the only child of $X_i$, the posterior probability distribution of $X_i$ is straightforward to compute exactly. Since the focus of the current paper is on

---

*Input:* Initialized importance function $\Pr^0(\mathbf{X}\backslash\mathbf{E})$, learning rate $\eta(k)$.
*Output:* An estimated importance function $\Pr^S(\mathbf{X}\backslash\mathbf{E})$.
**for** stage $k \leftarrow 0$ **to** $S$ **do**

1. Sample $l$ points $\mathbf{s}_1^k$, $\mathbf{s}_2^k$, ..., $\mathbf{s}_l^k$ independently according to the current importance function $\Pr^k(\mathbf{X}\backslash\mathbf{E})$.

2. For every node $X_i$ such that $X_i \in \mathbf{X}\backslash\mathbf{E}$ and $X_i \notin \mathrm{Anc}(\mathbf{E})$ count score sums corresponding to $\{x_i, \mathrm{pa}(X_i), \mathbf{e}\}$ and estimate $\Pr'(x_i|\mathrm{pa}(X_i), \mathbf{e})$ based on $\mathbf{s}_1^k$, $\mathbf{s}_2^k$, ..., $\mathbf{s}_l^k$.

3. Update $\Pr^k(\mathbf{X}\backslash\mathbf{E})$ according to the following formula:

$$\Pr^{k+1}(x_i|\mathrm{pa}(X_i), \mathbf{e}) = $$
$$\Pr^k(x_i|\mathrm{pa}(X_i), \mathbf{e}) + \eta(k) \cdot \Big(\Pr'(x_i|\mathrm{pa}(X_i), \mathbf{e}) - \Pr^k(x_i|\mathrm{pa}(X_i), \mathbf{e})\Big)$$

**end for**

---

Figure 3: The AIS-BN algorithm for learning the optimal importance function.

sampling, the test results reported in this paper do not include this improvement of the AIS-BN algorithm.

Figure 3 lists an algorithm that implements Step 7 of the basic AIS-BN algorithm listed in Figure 2. When we estimate $\Pr'(x_i|\text{pa}(X_i), \mathbf{e})$, we only use the samples obtained at the current stage. One reason for this is that the information obtained in previous stages has been absorbed by $\Pr^k(\mathbf{X}\backslash\mathbf{E})$. The other reason is that in principle, each successive iteration is more accurate than the previous one and the importance function is closer to the optimal importance function. Thus, the samples generated by $\Pr^{k+1}(\mathbf{X}\backslash\mathbf{E})$ are better than those generated by $\Pr^k(\mathbf{X}\backslash\mathbf{E})$. $\Pr'(X_i|\text{pa}(X_i), \mathbf{e}) - \Pr^k(X_i|\text{pa}(X_i), \mathbf{e})$ corresponds to the vector of first partial derivatives in the direction of the maximum decrease in the error. $\eta(k)$ is a positive function that determines the *learning rate*. When $\eta(k) = 0$ (lower bound), we do not update our importance function. When $\eta(k) = 1$ (upper bound), at each stage we discard the old function. The convergence speed is directly related to $\eta(k)$. If it is small, the convergence will be very slow due to the large number of updating steps needed to reach a local minimum. On the other hand, if it is large, convergence rate will be initially very fast, but the algorithm will eventually start to oscillate and thus may not reach a minimum. There are many papers in the field of neural network learning that discuss how to choose the learning rate and let estimated importance function converge quickly to the destination function. Any method that can improve learning rate should be applicable to this algorithm. Currently, we use the following function proposed by Ritter et al. (1991)

$$\eta(k) = a \left(\frac{b}{a}\right)^{k/k_{\max}}, \tag{10}$$

where $a$ is the initial learning rate and $b$ is the learning rate in the last step. This function has been reported to perform well in neural network learning (Ritter et al., 1991).

### 3.3 Heuristic Initialization in AIS-BN

The dimensionality of the problem of Bayesian network inference is equal to the number of variables in a network, which in the networks considered in this paper can be very high. As a result, the learning space of the optimal importance function is very large. Choice of the initial importance function $\Pr^0(\mathbf{X}\backslash\mathbf{E})$ is an important factor affecting the learning — an initial value of the importance function that is close to the optimal importance function can greatly affect the speed of convergence. In this section, we present two heuristics that help to achieve this goal.

Due to their explicit encoding of a decomposable joint probability distribution, Bayesian networks offer computational advantages compared to finite-dimensional integrals. A possible first approximation of the optimal importance function is the prior probability distribution over the network variables, $\Pr(\mathbf{X})$. We propose an improvement on this initialization. We know that the effect of evidence nodes on a node will be attenuated as the path length of that node to evidence nodes is increased (Henrion, 1989) and the most affected nodes are the direct ancestors of the evidence nodes. Initializing the ICPT tables of the parents of the evidence nodes to uniform distributions in our experience improves the convergence rate. Furthermore, the CPT tables of the parents of an evidence node $E$ may be not favorable to the observed state $e$ if the probability of $E = e$ without any condition is less than a small value, such as $\Pr(E = e) < 1/(2 \cdot n_E)$, where $n_E$ is the number of outcomes of node $E$. Based on this observation, we change the CPT tables of the parents of an evidence node $E$ to uniform distributions in our experiment only when $\Pr(E = e) < 1/(2 \cdot n_E)$, otherwise we leave them unchanged. This kind of initialization involves the knowledge of $\Pr(E = e)$, the marginal probability without evidence. Probabilistic logic sampling (Henrion, 1988) enhanced by Latin hypercube sampling (Cheng & Druzdzel, 2000b) or quasi-Monte Carlo methods (Cheng & Druzdzel, 2000a) will produce a very good estimate of $\Pr(E = e)$. This is an one-time effort that can be made at the model building stage and is worth pursuing to any desired precision.

Another serious problem related to sampling are extremely small probabilities. Suppose there exists a root node with a state $s$ that has the prior probability $\Pr(s) = 0.0001$. Let the posterior probability of this state given evidence be $\Pr(s|\mathbf{E}) = 0.8$. A simple calculation shows that if we update the importance function every 1,000 samples, we can expect to hit $s$ only once every 10 updates. Thus $s$'s convergence rate will be very slow. We can overcome this problem by setting a threshold $\theta$ and replacing every probability $p < \theta$ in the network by $\theta$.[2] At the same time, we subtract $(\theta - p)$ from the largest probability in the same conditional probability distribution. For example, the value of $\theta = 10/l$, where $l$ is the updating interval, will allow us to sample 10 times more often in the first stage of the algorithm. If this state turns out to be more likely (having a large weight), we can increase its probability even more in order to converge to the correct answer faster. Considering that we should avoid $f(\mathbf{X}) \ll |g(\mathbf{X}) - I \cdot f(\mathbf{X})|$ in an unimportant region as discussed in Section 2.1, we need to make this threshold larger. We have found that the convergence rate is quite sensitive to this threshold. Based on our empirical tests, we suggest to use $\theta = 0.04$ in networks whose maximum number of outcomes per node does not exceed five. A smaller threshold might lead to fast convergence in some cases but slow convergence in others. If one threshold does not work, changing it in a specific network will usually improve convergence rate.

### 3.4 Selection of Parameters

There are several tunable parameters in the AIS-BN algorithm. We base the choice of these parameters on the Central Limit Theorem (CLT). According to CLT, if $Z_1, Z_2, \ldots, Z_n$ are independent and identically distributed random variables with $E(Z_i) = \mu_Z$ and $\text{Var}(Z_i) = \sigma_Z^2$, $i = 1, \ldots, n$, then $\overline{Z} = (Z_1 + \ldots + Z_n)/n$ is approximately normally distributed when $n$ is sufficiently large. Thus,

$$\lim_{n \to \infty} P\left(\frac{\left|\overline{Z} - \mu_z\right|}{\mu_z} \geq \frac{\sigma_Z/\sqrt{n}}{\mu_z} \cdot t\right) = \frac{2}{\sqrt{2\pi}} \int_t^\infty e^{-x^2/2} dx \ . \tag{11}$$

Although this approximation holds when $n$ approaches infinity, CLT is known to be very robust and lead to excellent approximations even for small $n$. The formula of Equation 11 is an $(\varepsilon_r, \delta)$ Relative Approximation, which is an estimate $\overline{\mu}$ of $\mu$ that satisfies

$$P\left(\frac{|\overline{\mu} - \mu|}{\mu} \geq \varepsilon_r\right) \leq \delta \ .$$

---

2. This initialization heuristic was apparently developed independently by Ortiz and Kaelbling (2000).

If $\delta$ has been fixed,

$$\varepsilon_r = \frac{\sigma_Z/\sqrt{n}}{\mu_z} \cdot \Phi_Z^{-1}(\frac{\delta}{2}) \, ,$$

where $\Phi_Z(z) = \frac{1}{\sqrt{2\pi}} \int_z^\infty e^{-x^2/2} dx$. Since in our sampling problem, $\mu_z$ (corresponding to $\Pr(\mathbf{E})$ in Figure 2) has been fixed, setting $\varepsilon_r$ to a smaller value amounts to letting $\sigma_Z/\sqrt{n}$ be smaller. So, we can adjust the parameters based on $\sigma_Z/\sqrt{n}$, which can be estimated using Equation 3. It is also the theoretical intuition behind our recommendation $w^k \propto 1/\hat{\sigma}^k$ in Section 3.1. While we expect that this should work well in most networks, no guarantees can be given here — there exist always some extreme cases in sampling algorithms in which no good estimate of variance can be obtained.

### 3.5 A Generalization of AIS-BN: The Problem of Estimating $\Pr(\mathbf{a}|\mathbf{e})$

A typical focus of systems based on Bayesian networks is the posterior probability of various outcomes of individual variables given evidence, $\Pr(a|\mathbf{e})$. This can be generalized to the computation of the posterior probability of a particular instantiation of a set of variables given evidence, i.e., $\Pr(\mathbf{A} = \mathbf{a}|\mathbf{e})$. There are two methods that are capable of performing this computation. The first method is very efficient at the expense of precision. The second method is less efficient, but offers in general better convergence rates. Both methods are based on Equation 7.

The first method reuses the samples generated to estimate $\Pr(\mathbf{e})$ in estimating $\Pr(\mathbf{a}, \mathbf{e})$. Estimation of $\Pr(\mathbf{a}, \mathbf{e})$ amounts to counting the scored sum under the condition $\mathbf{A} = \mathbf{a}$. The main advantage of this method is its efficiency — we can use the same set of samples to estimate the posterior probability of any state of a subset of the network given evidence. Its main disadvantage is that the variance of the estimated $\Pr(\mathbf{a}, \mathbf{e})$ can be large, especially when the numerical value of $\Pr(\mathbf{a}|\mathbf{e})$ is extreme. This method is the most widely used approach in the existing stochastic sampling algorithms.

The second method, used much more rarely (e.g., Cano et al., 1996; Pradhan & Dagum, 1996; Dagum & Luby, 1997), calls for estimating $\Pr(\mathbf{e})$ and $\Pr(\mathbf{a}, \mathbf{e})$ separately. After estimating $\Pr(\mathbf{e})$, an additional call to the algorithm is made for each instantiation $\mathbf{a}$ of the set of variables of interest $\mathbf{A}$. $\Pr(\mathbf{a}, \mathbf{e})$ is estimated by sampling the network with the set of observations $\mathbf{e}$ extended by $\mathbf{A} = \mathbf{a}$. The main advantage of this method is that it is much better at reducing variance than the first method. Its main disadvantage is the computational cost associated with sampling for possibly many combinations of states of nodes of interest.

Cano et al. (1996) suggested a modified version of the second method. Suppose that we are interested in the posterior distribution $\Pr(\mathbf{a}_i|\mathbf{e})$ for all possible values $\mathbf{a}_i$ of $\mathbf{A}$, $i = 1$, 2, ..., $k$. We can estimate $\Pr(\mathbf{a}_i, \mathbf{e})$ for each $i = 1$, ..., $k$ separately, and use the value $\sum_{i=1}^k \Pr(\mathbf{a}_i, \mathbf{e})$ as an estimate for $\Pr(\mathbf{e})$. The assumption behind this approach is that the estimate of $\Pr(\mathbf{e})$ will be very accurate because of the large sample from which it is drawn. However, even if we can guarantee small variance in every $\Pr(\mathbf{a}_i, \mathbf{e})$, we cannot guarantee that their sum will also have a small variance. So, in the AIS-BN algorithm we only use the pure form of each of the methods. The algorithm listed in Figure 2 is based on the first method when the optional computations in Steps 12 and 13 are performed. An algorithm corresponding to the second method skips the optional steps and calls the basic AIS-BN algorithm twice to estimate $\Pr(\mathbf{e})$ and $\Pr(\mathbf{a}, \mathbf{e})$ separately.

The first method is very attractive because of its simplicity and possible computational efficiency. However, as we have shown in Section 2.2, the performance of a sampling algorithm that uses just one set of samples (as in the first method above) to estimate $\Pr(\mathbf{a}|\mathbf{e})$ will deteriorate if the difference between the optimal importance functions for $\Pr(\mathbf{a},\mathbf{e})$ and $\Pr(\mathbf{e})$ is large. If the main focus of the computation is high accuracy of the posterior probability distribution of a small number of nodes, we strongly recommend to use the algorithm based on the second method. Also, this algorithm can be easily used to estimate confidence intervals of the solution.

## 4. Experimental Results

In this section, we first describe the experimental method used in our tests. Our tests focus on the CPCS network, which is one of the largest and most realistic networks available and for which we know precisely which nodes are observable. We were, therefore, able to generate very realistic test cases. Since the AIS-BN algorithm uses two initialization heuristics, we designed an experiment that studies the contribution of each of these two heuristics to the performance of the algorithm. To probe the extent of AIS-BN algorithm's excellent performance, we test it on several real and large networks.

### 4.1 Experimental Method

We performed empirical tests comparing the AIS-BN algorithm to the likelihood weighting (LW) and the self-importance sampling (SIS) algorithms. The two algorithms are basically the state of the art general purpose belief updating algorithms. The AA (Dagum et al., 1995) and the bounded variance (Dagum & Luby, 1997) algorithms, which were suggested by a reviewer, are essentially enhanced special purpose versions of the basic LW algorithm. Our implementation of the three algorithms relied on essentially the same code with separate functions only when the algorithms differed. It is fair to assume, therefore, that the observed differences are purely due to the theoretical differences among the algorithms and not due to the efficiency of implementation. In order to make the comparison of the AIS-BN algorithm to LW and SIS fair, we used the first method of computation (Section 3.5), i.e., one that relies on single sampling rather than calling the basic AIS-BN algorithm twice.

We measured the accuracy of approximation achieved by the simulation in terms of the Mean Square Error (*MSE*), i.e., square root of the sum of square differences between $\Pr'(x_{ij})$ and $\Pr(x_{ij})$, the sampled and the exact marginal probabilities of state $j$ $(j = 1, 2, \ldots, n_i)$ of node $i$, such that $X_i \notin \mathbf{E}$. More precisely,

$$MSE = \sqrt{\frac{1}{\sum_{X_i \in \mathbf{N} \backslash \mathbf{E}} n_i} \sum_{X_i \in \mathbf{N} \backslash \mathbf{E}} \sum_{j=1}^{n_i} (\Pr'(x_{ij}) - \Pr(x_{ij}))^2} \ ,$$

where $\mathbf{N}$ is the set of all nodes, $\mathbf{E}$ is the set of evidence nodes, and $n_i$ is the number of outcomes of node $i$. In all diagrams, the reported *MSE* is averaged over 10 runs. We used the clustering algorithm (Lauritzen & Spiegelhalter, 1988) to compute the gold standard results for our comparisons of the mean square error. We performed all experiments on a Pentium II, 333 MHz Windows computer.

While *MSE* is not perfect, it is the simplest way of capturing error that lends itself to further theoretical analysis. For example, it is possible to derive analytically the idealized convergence rate in terms of *MSE*, which, in turn, can be used to judge the quality of the algorithm. *MSE* has been used in virtually all previous tests of sampling algorithms, which allows interested readers to tie the current results to the past studies. A reviewer offered an interesting suggestion of using cross-entropy or some other technique that weights small changes near zero much more strongly than the equivalent size change in the middle of the $[0, 1]$ interval. Such measure would penalize the algorithm for imprecisions of possibly several orders of magnitude in very small probabilities. While this idea is interesting, we are not aware of any theoretical reasons as to why this measure would make a difference in comparisons between AIS-BN, LW and SIS algorithms. The *MSE*, as we mentioned above, will allow us to compare the empirically determined convergence rate to the theoretically derived ideal convergence rate. Theoretically, the *MSE* is inversely proportional to the square root of the sample size.

Since there are several tunable parameters used in the AIS-BN algorithm, we list the values of the parameters used in our test: $l = 2,500$; $w^k = 0$ for $k \leq 9$ and $w^k = 1$ otherwise. We stopped the updating process in Step 7 of Figure 2 after $k \geq 10$. In other words, we used only the samples collected in the last step of the algorithm. The learning parameters used in our algorithm are $k_{\max} = 10$, $a = 0.4$, and $b = 0.14$ (see Equation 10). We used an empirically determined value of the threshold $\theta = 0.04$ (Section 3.3). We only change the CPT tables of the parents of a special evidence node $A$ to uniform distributions when $\Pr(A = a) < 1/(2 \cdot n_A)$. Some of the parameters are a matter of design decision (e.g., the number of samples in our tests), others were chosen empirically. Although we have found that these parameters may have different optimal values for different Bayesian networks, we used the above values in all our tests of the AIS-BN algorithm described in this paper. Since the same set of parameters led to spectacular improvement in accuracy in all tested networks, it is fair to say that the superiority of the AIS-BN algorithm to the other algorithms is not too sensitive to the values of the parameters.

For the SIS algorithm, $w^k = 1$ by the design of the algorithm. We used $l = 2,500$. The updating function in Step 7 of Figure 1 is that of (Shwe et al., 1991; Cousins, Chen, & Frisse, 1993):

$$\Pr^k_{new}(x_i | \mathrm{pa}(X_i), \mathbf{e}) = \frac{\Pr(x_i | \mathrm{pa}(X_i)) + k \cdot \widehat{\Pr}_{current}(x_i | \mathrm{pa}(X_i), \mathbf{e})}{1 + k} \; ,$$

where $\Pr(x_i | \mathrm{pa}(X_i))$ is the original sampling distribution, $\widehat{\Pr}_{current}(x_i | \mathrm{pa}(X_i), \mathbf{e})$ is an equivalent of our ICPT tables estimator based on all currently available information, and $k$ is the updating step.

## 4.2 Results for the CPCS Network

The main network used in our tests is a subset of the CPCS (Computer-based Patient Case Study) model (Pradhan et al., 1994), a large multiply-connected multi-layer network consisting of 422 multi-valued nodes and covering a subset of the domain of internal medicine.

Among the 422 nodes, 14 nodes describe diseases, 33 nodes describe history and risk factors, and the remaining 375 nodes describe various findings related to the diseases. The CPCS network is among the largest real networks available to the research community at the present time. The CPCS network contains many extreme probabilities, typically on the order of $10^{-4}$. Our analysis is based on a subset of 179 nodes of the CPCS network, created by Max Henrion and Malcolm Pradhan. We used this smaller version in order to be able to compute the exact solution for the purpose of measuring approximation error in the sampling algorithms.

The AIS-BN algorithm has some learning overhead. The following comparison of execution time vs. number of samples may give the reader an idea of this overhead. Updating the CPCS network with 20 evidence nodes on our system takes the AIS-BN algorithm a total of 8.4 seconds to learn. It generates subsequently 3,640 samples per second, while the SIS algorithm generates 2,631 samples per second, and the LW algorithm generates 4,167 samples per second. In order to remain conservative towards the AIS-BN algorithm, in all our experiments we fixed the execution time of the algorithms (our limit was 60 seconds) rather than the number of samples. In the CPCS network with 20 evidence nodes, in 60 seconds, AIS-BN generates about 188,000 samples, SIS generates about 158,000 samples and LW generates about 250,000 samples.

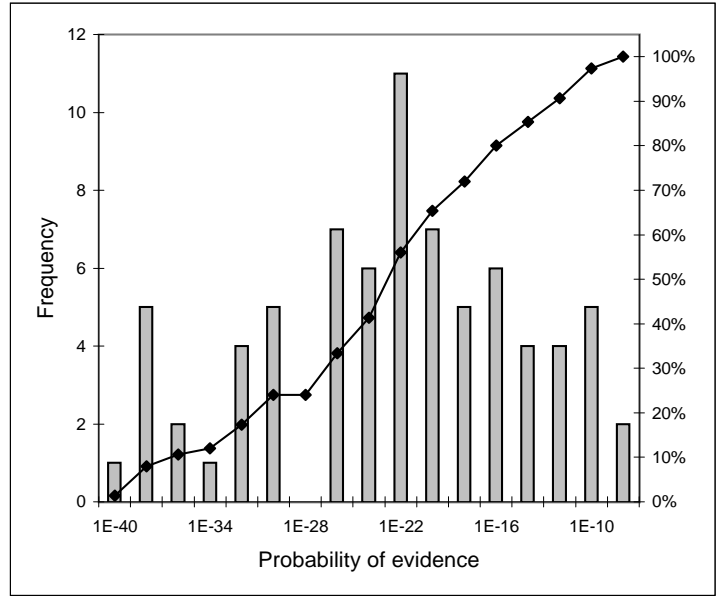

Figure 4: The probability distribution of evidence $\Pr(\mathbf{E} = \mathbf{e})$ in our experiments.

We generated a total of 75 test cases consisting of five sequences of 15 test cases each. We ran each test case 10 times, each time with a different setting of the random number seed. Each sequence had a progressively higher number of evidence nodes: 15, 20, 25, 30, and 35 evidence nodes respectively. The evidence nodes were chosen randomly (equiprobable sampling without replacement) from those nodes that described various plausible medical findings. Almost all of these nodes were leaf nodes in the network. We believe that this constituted very realistic test cases for the algorithms. The distribution of the prior probability of evidence, $\Pr(\mathbf{E} = \mathbf{e})$, across all test runs of our experiments is shown in Figure 4. The least likely evidence was $5.54 \times 10^{-42}$, the most likely evidence was $1.37 \times 10^{-9}$, and the median was $7 \times 10^{-24}$.

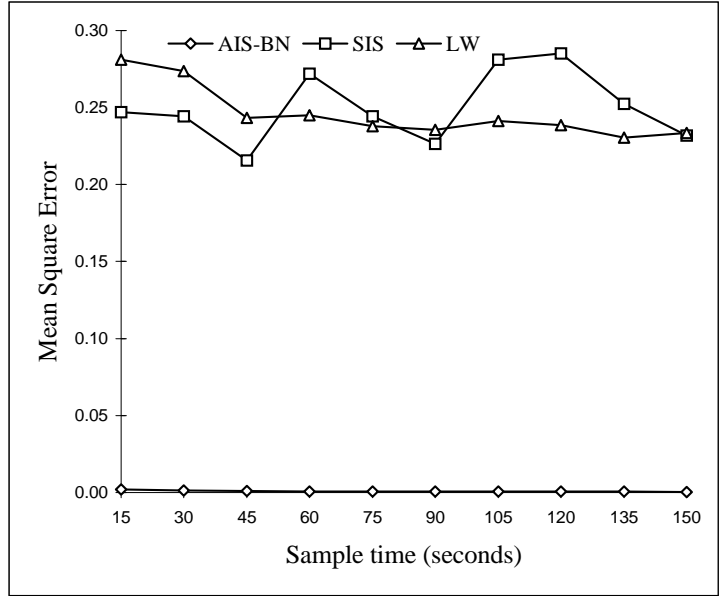

Figure 5: A typical plot of convergence of the tested sampling algorithms in our experiments — Mean Square Error as a function of the execution time for a subset of the CPCS network with 20 evidence nodes chosen randomly among plausible medical observations ($\Pr(\mathbf{E} = \mathbf{e}) = 3.33 \times 10^{-26}$ in this particular case) for the AIS-BN, the SIS, and the LW algorithms. The curve for the AIS-BN algorithm is very close to the horizontal axis.

Figures 5 and 6 show a typical plot of convergence of the tested sampling algorithms in our experiments. The case illustrated involves updating the CPCS network with 20 evidence nodes. We plot the *MSE* after the initial 15 seconds during which the algorithms start converging. In particular, the learning step of the AIS-BN algorithm is usually completed within the first 9 seconds. We ran the three algorithms in this case for 150 seconds rather than the 60 seconds in the actual experiment in order to be able to observe a wider range of convergence. The plot of the *MSE* for the AIS-BN algorithm almost touches the *X* axis in Figure 5. Figure 6 shows the same plot in a finer scale in order to show more detail in the AIS-BN convergence curve. It is clear that the AIS-BN algorithm dramatically improves the convergence rate. We can also see that the results of AIS-BN converge to exact results very fast as the sampling time increases. In the case captured in Figures 5 and 6, a tenfold increase in the sampling time (after subtracting the overhead for the AIS-BN algorithm, it corresponds to a 21.5-fold increase in the number of samples) results in a 4.55-fold decrease of the $MSE$ (to $MSE \leq 0.00048$). The observed convergence of both SIS and LW algorithms was poor. A tenfold increase in sampling time had practically no effect on accuracy. Please note that this is a very typical case observed in our experiments.

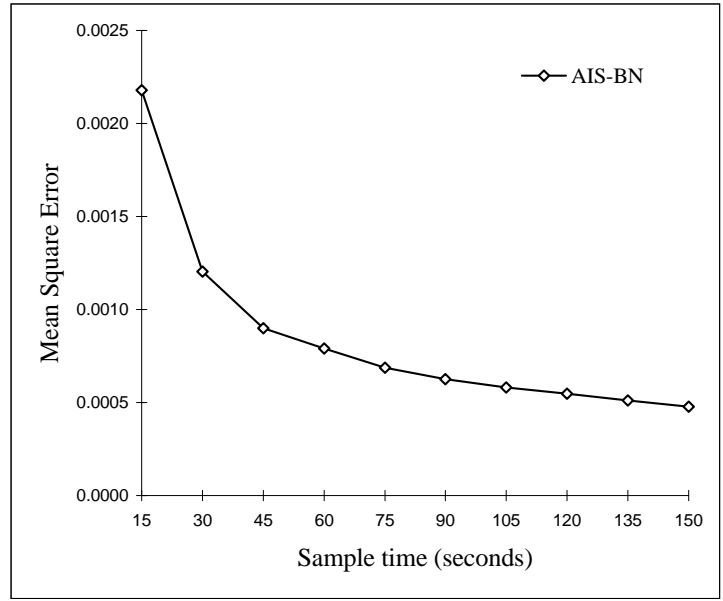

Figure 6: The lower part of the plot of Figure 5 showing the convergence of the AIS-BN algorithm to correct posterior probabilities.

| | Original CPT | Exact ICPT | Learned ICPT |
|---|---|---|---|
| "Absent" | 0.99631 | 0.0037 | 0.015 |
| "Mild" | 0.00183 | 0.1560 | 0.164 |
| "Moderate" | 0.00093 | 0.1190 | 0.131 |
| "Severe" | 0.00093 | 0.7213 | 0.690 |

Table 1: A fragment of the conditional probability table of a node of the CPCS network (node *gasAcute*, parents *hepAcute=Mild* and *wbcTotTho=False*) in Figure 6.

Figure 7 illustrates the ICPT learning process of the AIS-BN algorithm for the sample case shown in Figure 6. The displayed conditional probabilities belong to the node *gasAcute* which is a parent of two evidence nodes, *difInfGasMuc* and *abdPaiExaMea*. The node *gasAcute* has four states: "absent," "mild," "moderate," and "severe", and two parents. We randomly chose a combination of its parents' states as our displayed configuration. The original CPT for this configuration without evidence, the exact ICPT with evidence and the learned ICPT with evidence are summarized numerically in Table 1. Figure 7 illustrates that the learned importance conditional probabilities begin to converge to the exact results stably after three updating steps. The learned probabilities in Step 10 are close to the exact results. In this example, the difference between $\Pr(x_i|\mathrm{pa}(X_i), \mathbf{e})$ and $\Pr(x_i|\mathrm{pa}(X_i))$ is very large. Sampling from $\Pr(x_i|\mathrm{pa}(X_i))$ instead of $\Pr(x_i|\mathrm{pa}(X_i), \mathbf{e})$ would introduce large variance into our results.

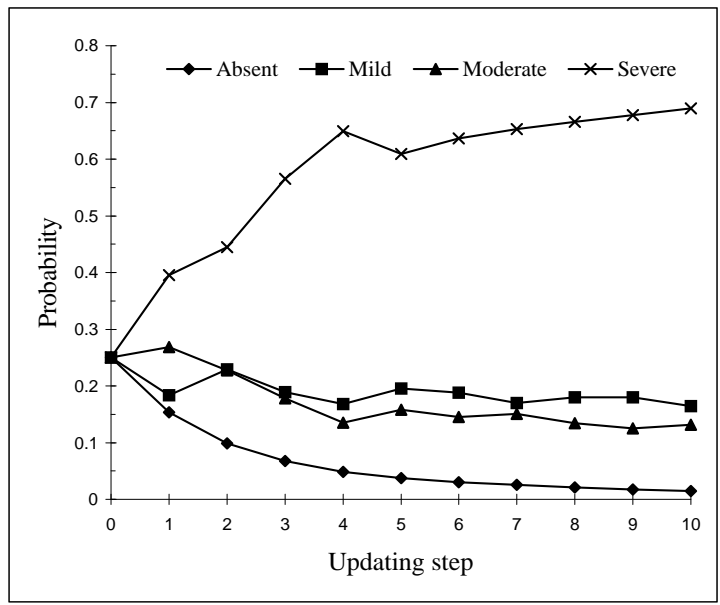

Figure 7: Convergence of the conditional probabilities during the example run of the AIS-BN algorithm captured in Figure 6. The displayed fragment of the conditional probability table belongs to node *gasAcute* which is a parent of one of the evidence nodes.

|        | AIS-BN  | SIS    | LW     |
|-------:|---------|--------|--------|
| $\mu$      | 0.00082 | 0.110  | 0.148  |
| $\sigma$   | 0.00022 | 0.076  | 0.093  |
| min    | 0.00049 | 0.0016 | 0.0031 |
| median | 0.00078 | 0.105  | 0.154  |
| max    | 0.00184 | 0.316  | 0.343  |

Table 2: Summary of the simulation results for all of the 75 simulation cases on the CPCS network. Figure 8 shows each of the 75 cases graphically.

Figure 8 shows the *MSE* for all 75 test cases in our experiments with the summary statistics in Table 2. A paired one-tailed *t*-test resulted in statistically highly significant differences between the AIS-BN and SIS algorithms ($p < 3.1 \times 10^{-20}$), and also between the SIS and LW algorithms ($p < 1.7 \times 10^{-8}$). As far as the magnitude of difference is concerned, AIS-BN was significantly better than SIS. SIS was better than LW, but the difference was small. The mean *MSE*s of SIS and LW algorithms were both greater than 0.1, which suggests that neither of these algorithms is suitable for large Bayesian networks.

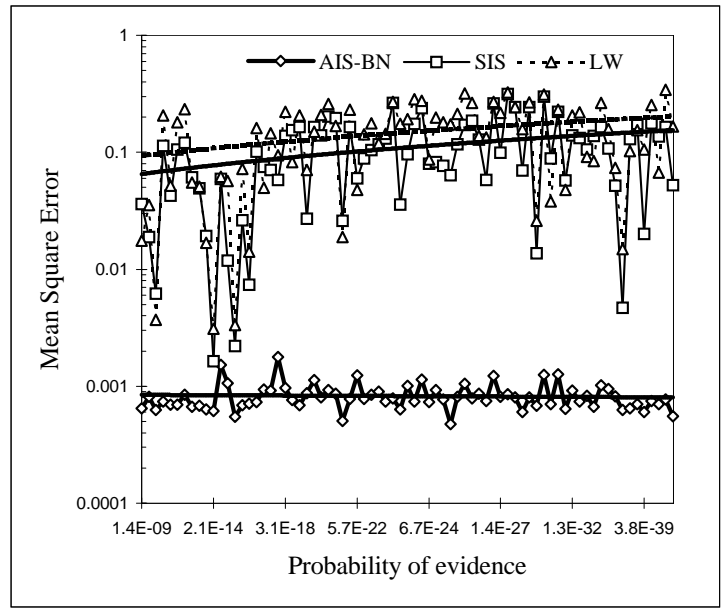

Figure 8: Performance of the AIS-BN, SIS, and LW algorithms: Mean Square Error for each of the 75 individual test cases plotted against the probability of evidence. The sampling time is 60 seconds.

The graph in Figure 9 shows the *MSE* ratio between the AIS-BN and SIS algorithms. We can see that the percentage of the cases whose ratio was greater than 100 (two orders of magnitude improvement!) is 60%. In other words, we obtained two orders of magnitude improvement in *MSE* in more than half of the cases. In 80% cases, the ratio was greater than 50. The smallest ratio in our experiments was 2.67, which happened when posterior probabilities were dominated by the prior probabilities. In that case, even though the LW and SIS algorithms converged very fast, their *MSE* was still far larger than that of AIS-BN.

Our next experiment aimed at showing how close the AIS-BN algorithm can approach the best possible sampling results. If we know the optimal importance sampling function, the convergence of the AIS-BN algorithm should be the same as that of forward sampling without evidence. In other words, the results of the probabilistic logic sampling algorithm without evidence approach the limit of how well stochastic sampling can perform. We ran the logic sampling algorithm on the CPCS network without evidence mimicking the test runs of the AIS-BN algorithm, i.e., 5 blocks of 15 runs, each repeated 10 times with a different random number seed. The number of samples generated was equal to the average number of samples generated by the AIS-BN algorithm for each series of 15 test runs. We obtained the average *MSE* $\mu = 0.00057$, with $\sigma = 0.000025$, min $= 0.00052$, and

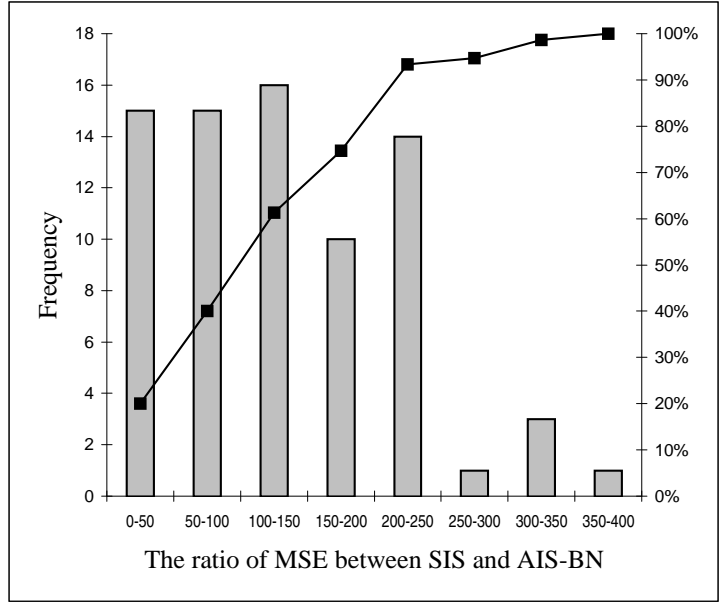

Figure 9: The ratio of *MSE* between SIS and AIS-BN versus percentage.

max = 0.00065. The best results should be around this range. From Table 2, we can see that the minimum *MSE* for the AIS-BN algorithm was 0.00049, within the range of the optimal result. The mean *MSE* in AIS-BN is 0.00082, not too far from the optimal results. The standard deviation, $\sigma$, is significantly larger in the AIS-BN algorithm, but this is understandable given that the process of learning the optimal importance function is heuristic in nature. It is not difficult to understand that there exist a difference between the AIS-BN results and the optimal results. First, the AIS-BN algorithm in our tests updated the sampling distribution only 10 times, which may be too few times to let it converge to the optimal importance distribution. Second, even if the algorithm has converged to the optimal importance distribution, the sampling algorithm will still let the parameter oscillate around this distribution and there will be always small differences between the two distributions.

Figure 10 shows the convergence rate for all tested cases for a four-fold increase in sampling time (between 15 and 60 seconds). We adjusted the convergence ratio of the AIS-BN algorithm by dividing it by a constant. According to Equation 3, the theoretically expected convergence ratio for a four-fold increase in the number of samples should be around two. There are about 96% cases among the AIS-BN runs whose ratio lays in the interval (1.75, 2.25], in a sharp contrast to 11% and 13% cases in the SIS and LW algorithms. The ratios of the remaining 4% cases in AIS-BN lay in the interval [2.25, 2.5]. In the SIS and LW algorithms, the percentage of cases whose ratio were smaller than 1.5 was 71% and 77% respectively. Less than 1.5 means that the number of samples was too small to estimate variance and the results cannot be trusted. The ratio greater than 2.25 means possibly that 60 seconds was long enough to estimate the variance, but 15 seconds was too short.

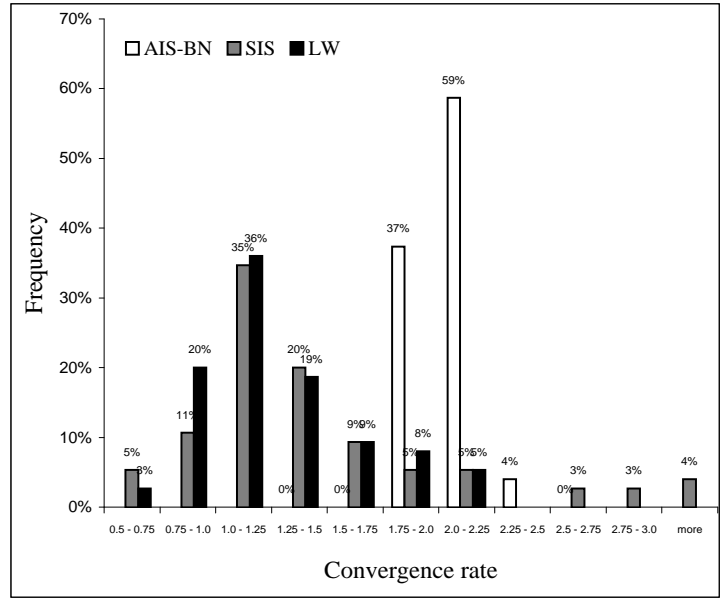

Figure 10: The distribution of the convergence ratio of the AIS-BN, SIS, and LW algorithms when the number of samples increases four times.

### 4.3 The Role of AIS-BN Heuristics in Performance Improvement

From the above experimental results we can see that the AIS-BN algorithm can improve the sampling performance significantly. Our next series of tests focused on studying the role of the two AIS-BN initialization heuristics. The first is initializing the ICPT tables of the parents of evidence to uniform distributions, denoted by $U$. The second is adjusting small probabilities, denoted by $S$. We denote AIS-BN without any heuristic initialization method to be the AIS algorithm. AIS+$U$+$S$ equals AIS-BN. We compared the following versions of the algorithms: SIS, AIS, SIS+$U$, AIS+$U$, SIS+$S$, AIS+$S$, SIS+$U$+$S$, AIS+$U$+$S$. All algorithms with SIS used the same number of samples as SIS. All algorithms with AIS used the same number of samples as AIS-BN. We tested these algorithms on the same 75 test cases used in the previous experiment. Figure 11 shows the *MSE* for each of the sampling algorithms with the summary statistics in Table 3. Even though the AIS algorithm is better than the SIS algorithm, the difference is not as large as in case of the AIS+$U$, AIS+$S$, and AIS-BN algorithms. It seems that heuristic initialization methods help much. The results for the SIS+$S$, SIS+$U$, SIS+$U$+$S$ algorithms suggest that although heuristic initialization methods can improve performance, they alone cannot improve too much. It is fair to say that significant performance improvement in the AIS-BN algorithm is coming from the combination of AIS with heuristic methods, not any method alone. It is not difficult to understand that, as only with good heuristic initialization methods is it possible to let the learning process quickly exit oscillation areas. Although both $S$ and $U$ methods alone can improve the performance, the improvement is moderate compared to the combination of the two.

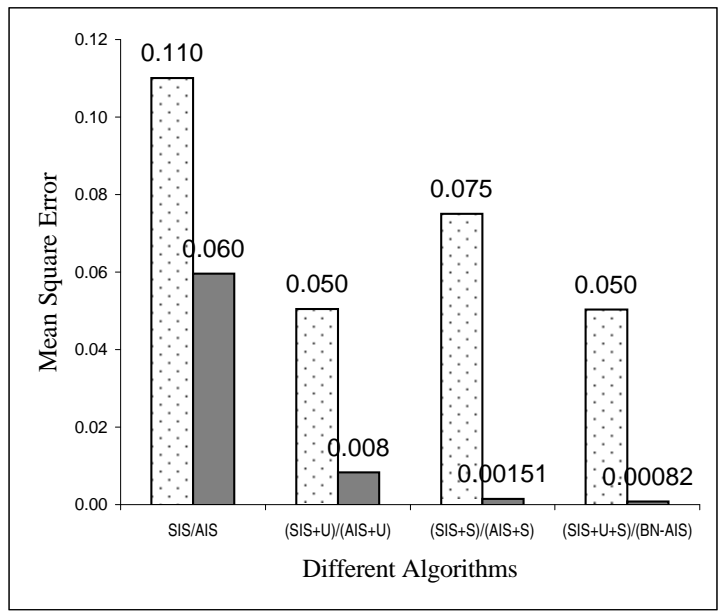

Figure 11: A comparison of different algorithms in the CPCS network. Each bar is based on 75 test cases. The dotted bar shows the *MSE* for the SIS algorithm while the gray bar shows the *MSE* for the AIS algorithm.

|        | SIS    | AIS     | SIS+U  | AIS+U   | SIS+S   | AIS+S   | SIS+U+S | AIS-BN  |
|--------|--------|---------|--------|---------|---------|---------|---------|---------|
| $\mu$    | 0.110  | 0.060   | 0.050  | 0.0084  | 0.075   | 0.0015  | 0.050   | 0.00082 |
| $\sigma$ | 0.076  | 0.049   | 0.052  | 0.025   | 0.074   | 0.0016  | 0.059   | 0.00022 |
| min    | 0.0016 | 0.00074 | 0.0011 | 0.00058 | 0.00072 | 0.00056 | 0.00086 | 0.00049 |
| median | 0.105  | 0.045   | 0.031  | 0.0014  | 0.052   | 0.00087 | 0.028   | 0.00078 |
| max    | 0.316  | 0.207   | 0.212  | 0.208   | 0.279   | 0.0085  | 0.265   | 0.0018  |

Table 3: Summary of the simulation results for different algorithms in the CPCS network.

## 4.4 Results for Other Networks

In order to make sure that the AIS-BN algorithm performs well in general, we tested it on two other large networks.

The first network that we used in our tests is the PATHFINDER network (Heckerman et al., 1990), which is the core element of an expert system that assists surgical pathologists with the diagnosis of lymph-node diseases. There are two versions of this network. We used the larger version, consisting of 135 nodes. In contrast to the CPCS network, PATHFINDER contains many conditional probabilities that are equal to 1, which reflects deterministic relationships in certain settings. To make the sampling challenging, we randomly selected 20 evidence nodes from among the leaf nodes. Each of these was an observable node (David Heckerman, personal communication). We verified in each case that the probability of so selected evidence was not equal to zero.

We fixed the execution time of the algorithms to be 60 seconds. The learning overhead for the AIS-BN algorithm in the PATHFINDER network was about 3.5 seconds. In 60 seconds, AIS-BN generated about 366,000 samples, SIS generated about 250,000 samples and LW generated about 2,700,000 samples. The reason why LW could generate more than 10 times as many samples as SIS within the same amount of time is that the LW algorithm terminates sample generation at a very early stage in many samples, when the weight of a sample becomes zero. This is a result of determinism in the probability tables, mentioned above. We will see that LW benefits greatly from generating more samples. The other parameters used in AIS-BN were the same as those used in the CPCS network.

We tested 20 cases, each with randomly selected 20 evidence nodes. The reported *MSE* for each case was averaged over 10 runs. Some of the runs of the SIS and LW algorithms did not manage to generate any effective samples (the weight score sum was equal to zero). SIS had only 75% effective runs and LW had only 89% effective runs, which means that in some runs SIS and LW were unable to yield any information about the posterior distributions. In all those cases, we discarded the run and only averaged over the effective runs. All runs in the AIS-BN algorithm were effective. We report our experimental results with the summary statistics in Table 4. From these data, we can see that the AIS-BN algorithm is still significantly better than the SIS and LW algorithms. Since the LW algorithm can generate more than ten times the number of samples than the SIS algorithm, its performance is better than that of the SIS algorithm.

|  | AIS-BN | SIS | LW |
|---|---|---|---|
| $\mu$ | 0.00050 | 0.166 | 0.089 |
| $\sigma$ | 0.00037 | 0.107 | 0.0707 |
| min | 0.00025 | 0.00116 | 0.00080 |
| median | 0.00037 | 0.184 | 0.0866 |
| max | 0.0017 | 0.467 | 0.294 |
| effective runs | 200 | 150 | 178 |

Table 4: Summary of the simulation results for all of the 20 simulation cases on the PATHFINDER network.

The second network that we tested was one of the ANDES networks (Conati et al., 1997). ANDES is an intelligent tutoring system for classical Newtonian physics that is being developed by a team of researchers at the Learning Research and Development Center at the University of Pittsburgh and researchers at the United States Naval Academy. The student model in ANDES uses a Bayesian network to do long–term knowledge assessment, plan recognition, and prediction of students' actions during problem solving. We selected the largest ANDES network that was available to us, consisting of 223 nodes.

In contrast to the previous two networks, the depth of the ANDES network was significantly larger and so was its connectivity. There were only 22 leaf nodes. It is quite predictable that this kind of networks will pose difficulties to learning. We selected 20 evidence nodes randomly from the potential evidence nodes and tested 20 cases. All parameters were the same as those used in the CPCS network. We fixed the execution time of the algorithms to be 60 seconds. The learning overhead for the AIS-BN algorithm in the ANDES network was 13.4 seconds. In 60 seconds, AIS-BN generated about 114,000 samples, SIS generated about 98,000 samples and LW generated about 180,000 samples. In this network, LW still can generate almost two times the number of samples generated by the SIS algorithm.

We report our experimental results with the summary statistics in Table 5. The results show that also in the ANDES network the AIS-BN algorithm was significantly better than the SIS and LW algorithms. Since LW generated almost two times the number of samples that were generated by the SIS algorithm, its performance was better than that of the SIS algorithm.

|        | AIS-BN | SIS    | LW     |
|-------:|--------|--------|--------|
| $\mu$    | 0.0059 | 0.0628 | 0.0404 |
| $\sigma$ | 0.0049 | 0.102  | 0.0539 |
| min    | 0.0023 | 0.0028 | 0.0028 |
| median | 0.0045 | 0.0190 | 0.0198 |
| max    | 0.0237 | 0.321  | 0.221  |

Table 5: Summary of the simulation results for all of the 20 simulation cases on the ANDES network.

While the AIS-BN algorithm is on the average an order of magnitude more precise than the other two algorithms, the performance improvement is smaller than in the other two networks. The reason why the performance improvement of the AIS-BN algorithm over the SIS and LW algorithms in the ANDES network is smaller compared to that in the CPCS and PATHFINDER networks is that: (1) The ANDES network used in our tests was apparently not challenging enough for sampling algorithms in general. In the ANDES network, SIS and LW also can perform well in some cases. The minimum *MSE* of SIS and LW in our tested cases is almost the same as that of AIS-BN. (2) The number of samples generated by AIS-BN in this network is significantly smaller than that in the previous two networks and AIS-BN needs more time to learn. Although increasing the number of samples will improve the performance of all three algorithms, it improves the performance of AIS-BN more since the convergence ratio of the AIS-BN algorithm is usually larger than that of SIS and LW (see Figure 10). (3) The parameters that we used in this network were tuned for the CPCS network. (4) The large depth and fewer leaf nodes of the ANDES network pose some difficulties to learning.

## 5. Discussion

There is a fundamental trade-off in the AIS-BN algorithm between the time spent on learning the importance function and the time spent on sampling. Our current approach, which we believe to be reasonable, is to stop learning at the point when the importance function is good enough. In our experiments we stopped learning after 10 iterations.

There are several ways of improving the initialization of the conditional probability tables at the outset of the AIS-BN algorithm. In the current version of the algorithm, we initialize the ICPT table of every parent $N$ of an evidence node $E$ ($N \in \text{Pa}(E)$, $E \in \mathbf{E}$) to the uniform distribution when $\Pr(E = e) < 1/(2 \cdot n_E)$. This can be improved further. We can extend the initialization to those nodes that are severely affected by the evidence. They can be identified by examining the network structure and local CPTs.

We can view the learning process of the AIS-BN algorithm as a network rebuilding process. The algorithm constructs a new network whose structure is the same as the original network (except that we delete the evidence nodes and corresponding arcs). The constructed network models the joint probability distribution $\rho(\mathbf{X} \backslash \mathbf{E})$ in Equation 8, which approaches the optimal importance function. We use the learned $\rho'$ to approximate this distribution. If $\rho'$ approximates $\Pr(\mathbf{X}|\mathbf{E})$ accurately enough, we can use this new network to solve other approximate tasks, such as the problem of computing the Maximum A-Posterior assignment (MAP) (Pearl, 1988), finding $k$ most likely scenarios (Seroussi & Golmard, 1994), etc. A large advantage of this approach is that we can solve each of these problems as if the network had no evidence nodes.

We know that Markov blanket scoring can improve convergence rates in some sampling algorithms (Shwe & Cooper, 1991). It may also be applied to the AIS-BN algorithm to improve its convergence rate. According to Property 4 (Section 2.1), any technique that can reduce the variance $\sigma^2_{\Pr(\mathbf{e})}$ will reduce the variance of $\widehat{\Pr}(\mathbf{e})$ and correspondingly improve the sampling performance. Since the variance of stratified sampling (Rubinstein, 1981) is never much worse than that of random sampling, and can be much better, it can improve the convergence rate. We expect some other variance reduction methods in statistics, such as: ($i$) the expected value of a random variable; ($ii$) antithetic variants correlations (stratified sampling, Latin hypercube sampling, etc.); and ($iii$) systematic sampling, will also improve the sampling performance.

Current learning algorithm used a simple approach. Some heuristic learning methods, such as adjusting learning rates according to changes of the error (Jacobs, 1988), should also be applicable to our algorithm. There are several tunable parameters in the AIS-BN algorithm. Finding the optimal values of these parameters for any given network is another interesting research topic.

It is worth observing that the plots presented in Figure 8 are fairly flat. In other words, in our tests the convergence of the sampling algorithms did not depend too strongly on the probability of evidence. This seems to contradict the common belief that forward sampling schemes suffer from unlikely evidence. AIS-BN for one shows a fairly flat plot. The convergence of the SIS and LW algorithms seems to decrease slightly with unlikely evidence. It is possible that all three algorithms will perform much worse when the probability of evidence drops below some threshold value, which our tests failed to approach. Until this relationship has been studied carefully, we conjecture that the probability of evidence is not a good measure of difficulty of approximate inference.

Given that the problem of approximating probabilistic inference is NP-hard, there exist networks that will be challenging for any algorithm and we have no doubt that even the AIS-BN algorithm will perform poorly on them. To the day, we have not found such networks. There is one characteristic of networks that may be challenging to the AIS-BN algorithm. In general, when the number of parameters that need to be learned by the AIS-BN algorithm increases, its performance will deteriorate. Nodes with many parents, for example, are challenging to the AIS-BN learning algorithm, as it has to update the ICPT tables under all combinations of the parent nodes. It is possible that conditional probability distributions with causal independence properties, such as Noisy-OR distributions (Pearl, 1988; Henrion, 1989; Diez, 1993; Srinivas, 1993; Heckerman & Breese, 1994), common in very large practical networks, can be treated differently and lead to considerable savings in the learning time.

One direction of testing approximate algorithms, suggested to us by a reviewer, is to use very large networks for which exact solution cannot be computed at all. In this case, one can try to infer from the difference in variance at various stages of the algorithm whether it is converging or not. This is a very interesting idea that is worth exploring, especially when combined with theoretical work on stopping criteria in the line of the work of Dagum and Luby (1997).

## 6. Conclusion

Computational complexity remains a major problem in application of probability theory and decision theory in knowledge-based systems. It is important to develop schemes that improve the performance of updating algorithms — even though the theoretically demonstrated worst case will remain NP–hard, many practical cases may become tractable.

In this paper, we studied importance sampling in Bayesian networks. After reviewing the most important theoretical results related to importance sampling in finite-dimensional integrals, we proposed a new algorithm for importance sampling in Bayesian networks that we call *adaptive importance sampling* (AIS-BN). While the process of learning the optimal importance function for the AIS-BN algorithm is computationally intractable, based on the theory of importance sampling in finite-dimensional integrals we proposed several heuristics that seem to work very well in practice. We proposed heuristic methods for initializing the importance function that we have shown to accelerate the learning process, a smooth learning method for updating importance function using the structural advantages of Bayesian networks, and a dynamic weighting function for combining samples from different stages of the algorithm. All these methods help the AIS-BN algorithm to get fairly accurate estimates of the posterior probabilities in a limited time. Of the two applied heuristics, adjustment of small probabilities, seems to lead to the largest improvement in performance, although the largest decrease in *MSE* is achieved by a combination of the two heuristics with the AIS-BN algorithm.

The AIS-BN algorithm can lead to a dramatic improvement in the convergence rates in large Bayesian networks with evidence compared to the existing state of the art algorithms. We compared the performance of the AIS-BN algorithm to the performance of likelihood weighting and self-importance sampling on a large practical model, the CPCS network, with evidence as unlikely as $5.54 \times 10^{-42}$ and typically $7 \times 1.0^{-24}$. In our experiments, we observed that the AIS-BN algorithm was always better than likelihood weighting and self-importance sampling and in over 60% of the cases it reached over two orders of magnitude improvement in accuracy. Tests performed on the other two networks, PATHFINDER and ANDES, yielded similar results.

Although there may exist other approximate algorithms that will prove superior to AIS-BN in networks with special structure or distribution, the AIS-BN algorithm is simple and robust for general evidential reasoning problems in large multiply-connected Bayesian networks.

## Acknowledgments

We thank anonymous referees for several insightful comments that led to a substantial improvement of the paper. This research was supported by the National Science Foundation under Faculty Early Career Development (CAREER) Program, grant IRI–9624629, and by the Air Force Office of Scientific Research grants F49620–97–1–0225 and F49620–00–1–0112. An earlier version of this paper has received the 2000 School of Information Sciences Robert R. Korfhage Award, University of Pittsburgh. Malcolm Pradhan and Max Henrion of the Institute for Decision Systems Research shared with us the CPCS network with a kind permission from the developers of the INTERNIST system at the University of Pittsburgh. We thank David Heckerman for the PATHFINDER network and Abigail Gerner for the ANDES network used in our tests. All experimental data have been obtained using SMILE, a Bayesian inference engine developed at the Decision Systems Laboratory and available at `http://www2.sis.pitt.edu/~genie`.